\documentclass[letterpaper, 10 pt, conference]{ieeeconf}  

\IEEEoverridecommandlockouts                              

\usepackage{cite}
\usepackage{graphics} 
\usepackage{epsfig} 
\usepackage{amsmath}
\usepackage{units}
\usepackage{multirow}

\usepackage{footnote}
\makesavenoteenv{tabular}

\title{\LARGE \bf
An Inverting-Tube Clutching Contractile Soft Pneumatic Actuator}

\author{Wyatt~Felt
\thanks{Mechanical Engineering Department, \'{E}cole polytechnique f\'{e}d\'{e}rale de Lausanne (EPFL), 
        {\tt\small wfelt@umich.edu}
        }
}

\begin{document}

\maketitle
\thispagestyle{empty}
\pagestyle{empty}

\begin{abstract}

This paper presents the simple synergistic combination of a novel contracting soft pneumatic actuator with a soft clutch (linear brake). The device is designated the Inverting-tube Vacuum ACtuator with Clutch (InVACC). The actuator alone (no clutch) is designated ``InVAC'' and uses vacuum pressure to invert a thin tube into a shorter section of reinforced flexible tubing. The inverting tube acts as rolling diaphragm and a flexible tendon. This allows the actuator to contract to one third of its extended length. The contractile-force-per-unit-pressure is approximately constant over the stroke. The theoretical maximum of this force is the product of the vacuum gauge pressure and half the interior cross-sectional area of the tube. The experimental evaluation revealed hysteretic losses that depend on the actuation direction and rate. With \unit[-81]{kPa}, the prototype produced \unit[12.7]{N} of tension during extension and \unit[7.5]{N} during retraction. The reinforced tubing of the InVAC was integrated with an inner collapsible ``clutching'' tube to create an InVACC. The clutch is engaged by applying a positive pressure between the reinforced tube and the clutching tube, which collapses the clutching tube onto the flexible tendon. With a pressure of \unit[50]{kPa}, the InVACC clutch tested in this work was able to support a peak tensile load of \unit[120]{N} before slipping. Though the fatigue life of the current prototypes is limited, improved fabrication methods for this novel actuator/clutch concept will enable new applications in robotics and wearable haptic systems.

\end{abstract}

\section{INTRODUCTION}
Contracting Soft Pneumatic Actuators (SPAs) create muscle-like tensile forces from pressurized air.  Unlike rigid piston-cylinders, soft actuators have compliant structures and can contract even if their ends are not axially aligned. Soft contractile actuators are valued in biomimetics and wearable robotics for their light weight, compliance, low reflected-inertia and ability to create stable forces directly from input pressures.

Most types of contracting SPAs, however, are limited by their relatively short stroke lengths and nonlinear force output.  A McKibben muscle, for instance, has a contracted length that is, at its shortest, still approximately two-thirds of its extended length \cite{daerden_pneumatic_2002}. Other kinds of ``artificial muscles'' can reduce their length by about half \cite{daerden_pneumatic_2002}. Moreover, for the same internal pressure, the force output of these actuators changes dramatically over the course of their stroke. They commonly begin their contraction with a high force capability that drops to zero over the stroke \cite{Chou1996}.

\begin{figure}[!t]
\centering
\includegraphics[width=3.4in]{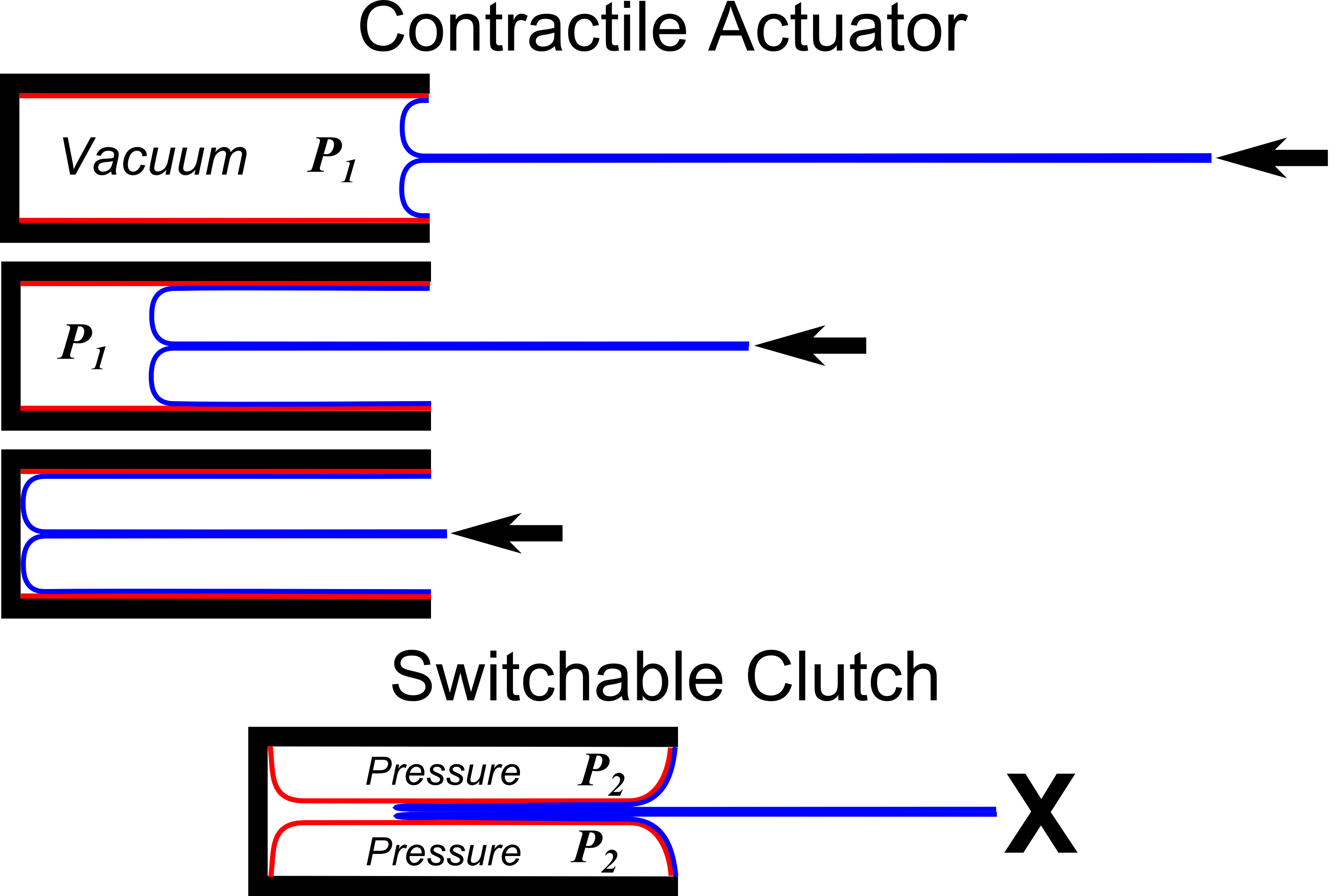}
\caption{The Inverting-tube Vacuum ACtuator with Clutch (InVACC) is a contractile linear actuator and clutch. It uses negative gauge pressure to draw a thin inverting tube (blue) into a reinforced flexible tube (black). This permits the actuator to contract to one-third of its extended length. The inverting tube acts as both a rolling diaphragm and a flexible tendon. The contraction force is constant and is controllable through a linear relationship with the applied vacuum pressure. The actuator (InVAC) can be combined with a clutch (InVACC). The clutch locks the actuator extension by collapsing an internal tube (red) onto the inverting tendon.}
\label{fig:system_description}
\end{figure}

These challenges are not universal to all SPAs. Extending McKibben muscles and bellows can extend to several multiples of their contracted length with a more constant force output \cite{hawkes_300_strain_2016}. Other extending actuators can extend almost indefinitely by unrolling spooled material \cite{hammond_pneumatic_2017,HawkesGrowth2017}. Some of these extending SPAs can be converted into contracting SPAs by applying vacuum pressures \cite{Belforte_textile,RodrigueVacuumBellows,UsevitchICRA2018,feltVacuumBellows} or by using elastic forces to recoil the actuators when pressure is removed \cite{hawkes_300_strain_2016,hammond_pneumatic_2017}. 

Linear clutches, or brakes, are also useful in engineered systems. These devices restrict the extension of flexible tendons when they are engaged. If the tension exceeds a certain level, the tendon may be able to extend, but its motion will be resisted with negative work from the clutch. Because they can resist motion by absorbing energy, these devices can be used as linear dampers \cite{choi_soft_2018}. They may also be able to stably render stiff haptic boundaries without high-bandwidth force control. Linear clutches have been used to store energy in wearable walking assistance devices \cite{collins_reducing_2015,diller_lightweight_2016} and have been proposed for use in safety restraints and muscle rehabilitation \cite{choi_soft_2018}.

Some previously proposed linear clutches rely on friction between overlapping sheets of material. By modulating the interface pressure between the sheets with either vacuum \cite{choi_soft_2018} or electrostatic \cite{diller_lightweight_2016} pressure, the frictional forces can be controlled.  The benefits of these other techniques include the ability to boost the effective surface area of the frictional clutch with multiple layers. The disadvantages include the limited interface pressure, the elasticity needed to draw the sheets together and the \unit[100]{\%} limit on extension strain.

\begin{figure}[!t]
\centering
\includegraphics[width=3in]{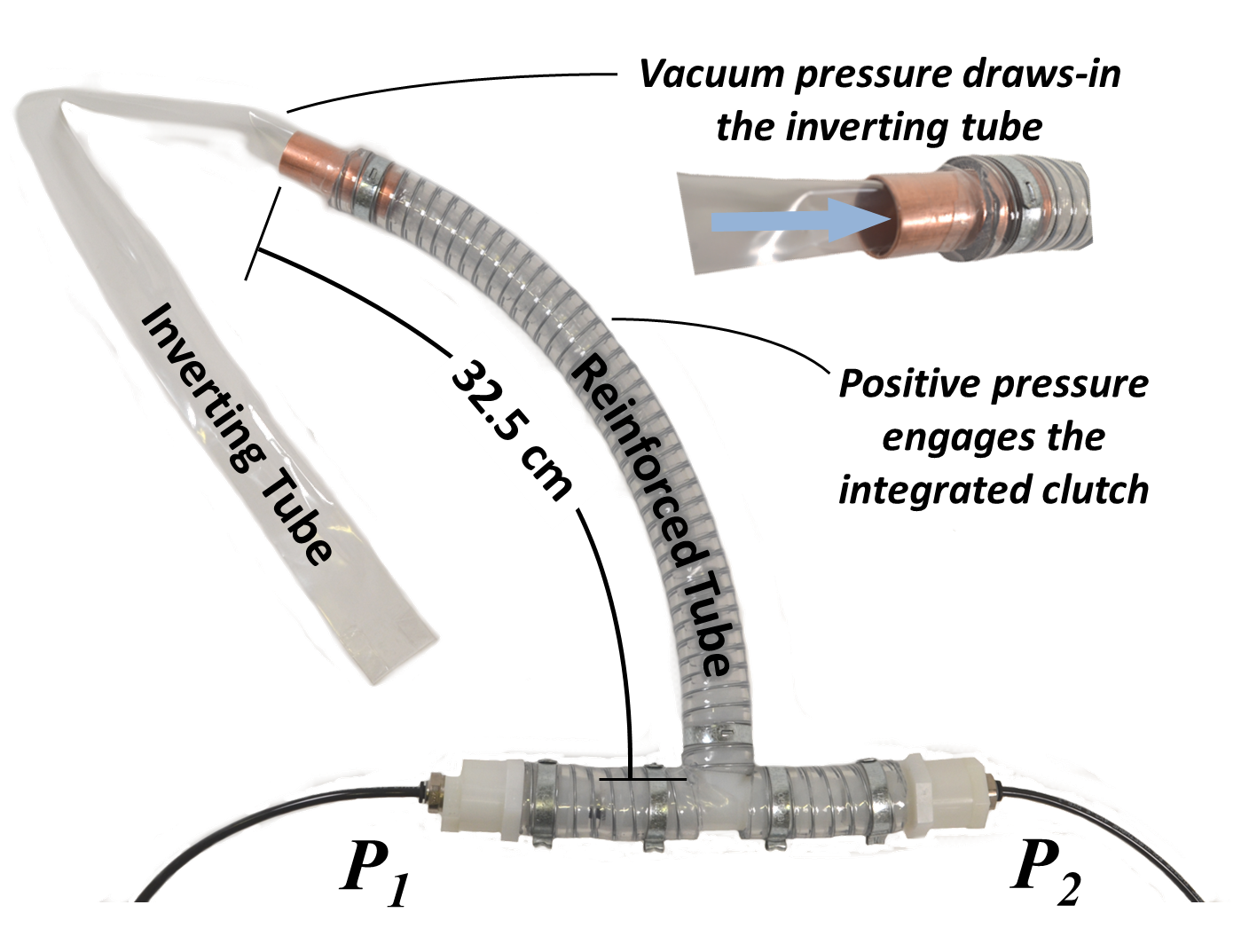}
\caption{The Inverting-tube Vacuum ACtuator with Clutch (InVACC) is a flexible linear actuator/clutch. It uses negative gauge pressure to draw a thin inverting tube into a thicker, reinforced tube. Pressurizing another thin tube inside of the reinforced tube engages a clutch that locks the extension of the device.}
\label{fig:proto}
\end{figure}

This work presents a soft contractile actuator, the Inverting-tube Vacuum ACtuator (InVAC, Figs.~\ref{fig:system_description} and \ref{fig:proto}), that is able to contract to a third of its extended length with a constant contractile-force-per-unit-pressure.  The actuator works by connecting the ends of a thin-walled ``inverting'' tube and a thick-walled ``reinforced'' tube. When the reinforced tube is subjected to negative internal gauge pressure (vacuum), the thin-walled tube inverts and retracts inside of the reinforced tube (Fig.~\ref{fig:system_description}, blue). This actuation mechanism is similar to the ``growing'' inverting-tube robot presented by Hawkes et al. \cite{HawkesGrowth2017}. For the InVAC, rather than locomoting with positive gauge pressure, the inverting tube is subjected to a negative gauge pressure to create a controllable tensile force. The inverting tube acts as a ``rolling diaphragm.'' Rolling diaphragm cylinders have been used for decades and were the subject of a recent investigation by Disney Research \cite{IROS_rolling_diaphram}.

The InVAC can be combined with an internal clutch. This combined device is designated as an Inverting-tube Vacuum ACtuator with Clutch (InVACC, Figs.~\ref{fig:proto} and \ref{fig:system_description}). The clutch uses positive pneumatic pressure to apply friction to the inverting tube. The pressure is applied between the reinforced tube and an additional thin tube that collapses onto the inverting tube (Fig.~\ref{fig:system_description}, red). The friction mechanism is similar to vacuum ``layer-jamming'' devices \cite{choi_soft_2018,kim_novel_2013,langer_stiffening_2018}. The InVACC, however, creates frictional pressure from positive gauge pressure, not vacuum. An additional advantage comes from the controllability of the actuation forces. Whereas other soft clutches rely on ever-present elastic forces to manage the slack in the clutch, the active tensile force from an InVACC can be switched on and off. This synergistic clutch/actuator combination results in a wide range of controllable forces--from zero to the material limits.

The contributions of this paper include the introduction of the novel InVAC inverting contractile actuator concept and the experimental evaluation of a physical prototype. This actuator enables a unique positive-pressure clutch (InVACC) that does not need elasticity for slack management. The paper presents the InVACC concept (\ref{sec:sys_description}),  models for the actuation and clutching (\ref{sec:act_model} and \ref{sec:clutch_model}), and an experimental characterization of the tensile forces (\ref{sec:exp_characterization}). The experimental results are presented and analyzed (\ref{sec:results}). This is followed by a discussion and conclusions (\ref{sec:disc}, which includes a comparison to other linear clutch technologies). The experimental results demonstrate the viability of the InVACC concept and reveal the limitations of its current embodiment. This new actuator/clutch concept will be a powerful tool for robotics and wearable haptic systems.

\section{METHODS}

\subsection{System Description}
\label{sec:sys_description}

The contractile force of the proposed actuator comes from a thin-walled inverting tube that acts both as a tendon and as a rolling diaphragm (Fig.~\ref{fig:system_description}, blue).
One end of the inverting tube is connected to a thicker-walled reinforced tube that is flexible but not collapsible.
When the reinforced tube is subjected to negative internal gauge pressure, the thin-walled tube inverts and retracts inside it.
The consistent rate of volume-change over the stroke makes the contractile force approximately constant for a constant retraction pressure $P_1$.
When fully contracted, the reinforced tube contains twice its length in inverting tubing.
This allows the actuator to contract to one third of its extended length.  

When partially contracted, the extension of the inverting tube can be restricted by engaging a clutch (Fig.~\ref{fig:system_description}, red). The clutching pressure $P_2$ is applied between the reinforced tube and a thin internal tube along its length. The pressure collapses the internal tube onto the inverting tube. This creates a frictional force that scales with the applied clutching pressure. 

The InVACC is able to operate in three distinct modes (Table~\ref{tab:modes}). By setting the retraction pressure ($P_1$) and the clutching pressure ($P_2$) to zero (i.e. the ambient pressure), the system can become completely inactive. In this mode, if the inverting tendon is extracted, it will not be actively retracted. This inactive mode could be useful, for example, in a wearable system. By deactivating the actuator/clutch, users could move unimpeded. In contrast, clutches that rely on elastic energy to retract need to be physically disconnected to stop applying spring-like restoring forces.

\begin{table}[h]
    \centering
    \caption{Actuation and Clutching Modes}
    \begin{tabular}{|c|c|c|c|c|}\hline
      \multirow{2}{*}{$P_1$}  & \multirow{2}{*}{$P_2$} & \multirow{ 2}{*}{Mode}   & Positive & Negative \\ 
                              &                        &                         &  Work    &   Work    \\ \hline \hline
        0    &  0    & No Tension (Device Inactive) & No & No \\ \hline
		-    &  -    & Controllable Tension & Yes & Yes \\ \hline
		-	 &  +    & Clutch Engaged & No & Yes \\ \hline
    \end{tabular}   
    \label{tab:modes}
\end{table}

When active, the tension generated by the InVACC can be controlled by modulating the vacuum pressure $P_1$. Pressurizing both chambers to negative gauge pressures ($P_1 < 0$, $P_2 < 0$) results in a controllable tensile force that scales with the magnitude of the pressure.  The resulting contractile forces could be used, for example, to render haptic information to a user while still enabling motion. They could also be used to manage the slack in the tendon in preparation for clutching.

The extension of the InVACC can be restricted by engaging the clutch with a positive pressure ($P_2 > 0$). The tension forces required to extend the actuator in this mode can be much higher than those created through the vacuum retraction. Importantly, however, the frictional forces from the clutch can only perform negative work. This makes the clutched mode useful in rendering inherently stable haptic boundaries. The passivity of the clutch avoids the instabilities that can result from attempting to render stiff haptic surfaces with low-bandwidth actuators or slow control-loops.

\begin{figure}[!t]
\centering
\includegraphics[width=2.5in]{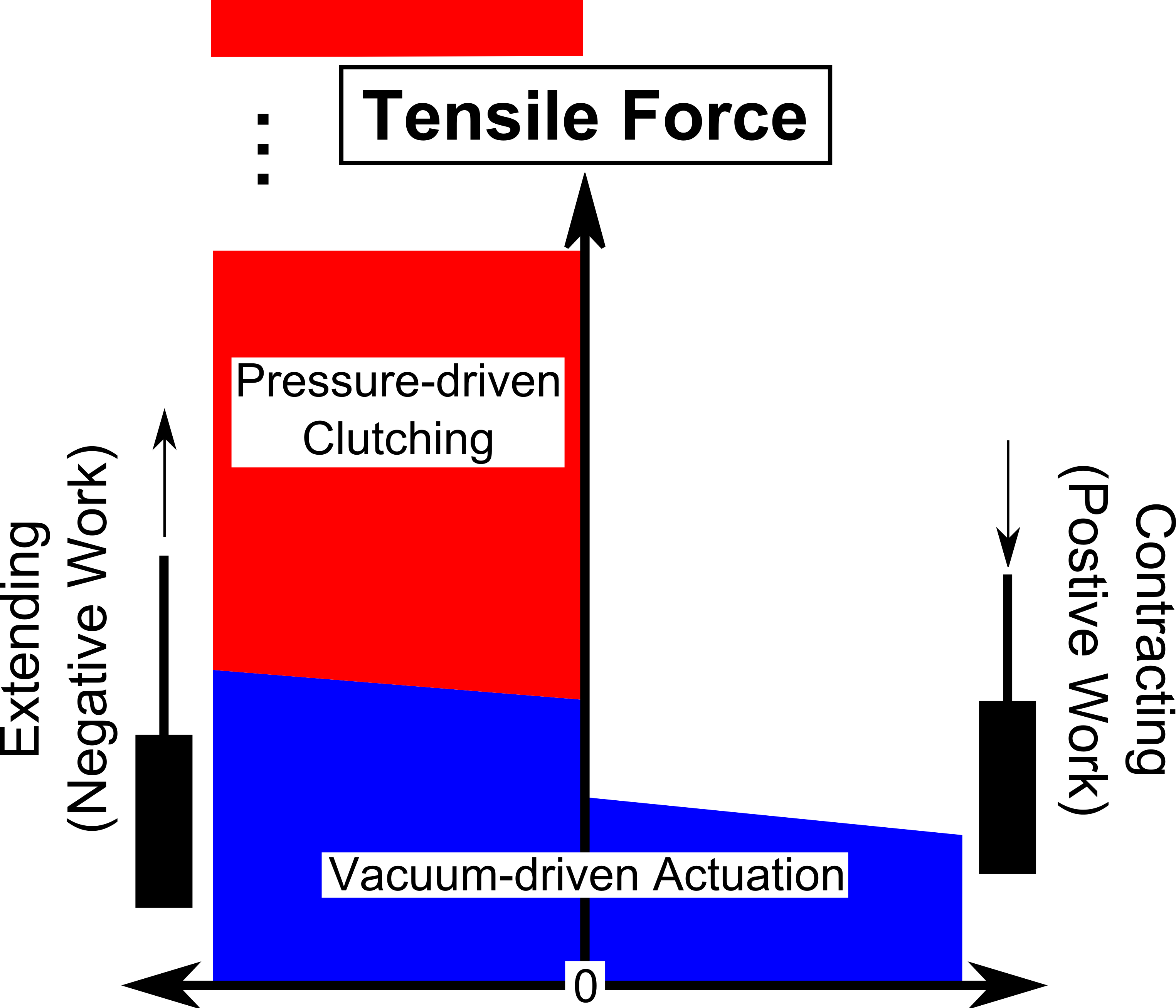}
\caption{The force capability of the InVACC combined actuator/clutch.  The internal vacuum pressure can be modulated to control the tensile actuation force (for both extension and contraction). The clutch pressure can be modulated to control the frictional forces that resist extension. Finally, the tension force can be reduced completely to zero by switching off the pressures and allowing slack to develop in the system.}
\label{fig:force_capability}
\end{figure}

The combination of these modes results in a large range of achievable forces (Fig.~\ref{fig:force_capability}). The vacuum pressures that can be applied to the device cannot exceed the ambient pressure. This caps the actuation forces that can be created from a device with a given internal diameter. Due to losses, the actuator requires more tension to extend (negative work) than it generates in contraction (positive work). If higher forces are needed to resist extension, the clutch can be used. This clutching force is limited only by the tensile strength of the inverting-tube tendon and the pressure limits of the other tubes.

\subsection{Contractile Actuator Model}
\label{sec:act_model}

The actuation forces come from the vacuum-driven reduction of the internal volume $V$ within the reinforced tube. The contractile force $F_{\text{act}}$ is reduced by losses in the system, characterized by $F_{\text{loss}}$. For an extension distance $x$ with a time rate-of-change $\dot{x}$, the actuation force can be approximated by 
\begin{equation}
\begin{aligned}
    F_{\text{act}} &= -\frac{dV}{dx}P_1 - F_{\text{loss}} \\
    & = -\frac{1}{2} \frac{\pi D^2}{4} P_1 - F_{\text{loss}}\!\left(\text{Sign}\left(\dot{x}\right),\left|\dot{x}\right|\right)
\end{aligned}
\label{eq:force_model_general}
\end{equation}
where $D$ is the maximum diameter of the inverting tube. Note that the actuation force scales with \textit{half} of the product of pressure and cross-sectional area (in contrast to a piston-cylinder actuator which scales with the full product). This difference comes from the fact that an inverting tube ``doubles back'' within the reinforced section. This results in a reduction in mechanical advantage similar to that from a single movable pulley. That is, the tendon extends twice as far as the change in the length of the empty portion of the reinforced tube.

The model used in this work assumes that the effective area $A_{\text{eff}}$ of the actuator may be slightly less than the ideal. This reduction could result, for example, from the bulk of the collapsed inverting-tube material. The loss force is assumed to consist of a yield force $F_{\text{yield}}$ required to deform the inverting tube and a viscous force that increases with the actuation rate and scales with the coefficient $\mu_{\text{visc}}$
\begin{equation}
    F =  -\frac{1}{2} A_{\text{eff}} P_1 - F_{\text{yield}}\text{Sign}\left(\dot{x}\right) - \mu_{\text{visc}}\dot{x}.
    \label{eq:force_model_specific}
\end{equation}

This model is similar to that used by Hawkes et al. in \cite{HawkesGrowth2017}. However, their model did not consider external forces or the corresponding halving of the pressure-area product. Moreover, the viscous losses proposed in \cite{HawkesGrowth2017} use a value of $\dot{x}$ raised to a power near unity (0.91). The model used in this work neglects other sources of loss identified for the comparatively long length system presented in \cite{HawkesGrowth2017} including the friction-per-unit-length and the friction from curvature.

Neglecting the losses, the maximum tensile force $F_{\text{max}}$ achievable by an InVAC is given by the inverting-tube diameter and the ambient pressure. At standard atmospheric pressures ($P_1 = -P_\text{atm} =$ \unitfrac[-101325]{N}{m$^2$}), Eq.~\eqref{eq:force_model_general} can be adapted to express the maximum force (in Newtons) in terms of the inverting-tube diameter $D_{\text{cm}}$ (in centimeters)
\begin{equation}
\begin{aligned}
    F_{\text{max}} &= -\frac{1}{2} \left(- P_\text{atm} \right) \frac{\pi D^2}{4} \\
    F_{\text{max}} &= 3.98 {D_{\text{cm}}}^2 .
\end{aligned}
\label{eq:force_max}
\end{equation}
Thus the maximum theoretical force capacity of the actuator at sea level is approximately \unit[$4$]{N} for every square centimeter of tube cross section. Some example theoretical maximum force values are shown in Table~\ref{tab:max_actuation_forces} for various values of the tube diameter.
\begin{table}[h]
    \centering
     \caption{Maximum actuation force at atmospheric pressure}
    \begin{tabular}{|c|c|}\hline
      Inverting Tube Diameter (cm)  & Max. Force (N) \\ \hline \hline
        0.25     &   0.25 \\ \hline
        0.5     &   1 \\ \hline
        1       &   4 \\ \hline
        2.5     &   25 \\ \hline
        5  &     100  \\ \hline
    \end{tabular}   
    \label{tab:max_actuation_forces}
\end{table}

\subsection{Clutch Model}
\label{sec:clutch_model}

Assuming a linearly elastic inverting tube material that does not elongate plastically, the force output from the clutch $F_{\text{clutch}}$ can be approximated as
\begin{equation}
    F_{\text{clutch}} =
    \begin{cases}
    0,              & x < x_{\text{clutch}} \\
    k_{\text{tube}}\left(x - x_{\text{clutch}}\right),  &  x_{\text{clutch}}  < x < x_{\text{slip}} \\ F_{\text{slip}}\!\left(x,\dot{x},P_1,P_2 \right), & x > x_{\text{slip}}, \: \dot{x} > 0 \\
\end{cases}
\label{eq:F_clutch}
\end{equation}
where $x$ is a measure of the tendon extension, $k_{\text{tube}}$ is a linear approximation of the stiffness (which may change for different levels of actuator extension), and $F_{\text{slip}}$ is the threshold above which the clutch will slip. The largest level of $x$ for which there is no clutch force is designated $x_{\text{clutch}}$. The extension at which the clutch will slip is designated $x_{\text{slip}}$. These levels are described by
\begin{equation}
    x_{\text{slip}} = F_{\text{slip}}/k_{\text{tube}} + x_{\text{clutch}}
\end{equation}
\begin{equation}
    x_{\text{clutch}} = x\!\left(t_{\text{clutch}}\right) + \int_{t_{\text{clutch}}}^{t} \begin{cases}
        \dot{x},         & x > x_{\text{slip}} \, \land \, \dot{x} > 0 \\
    0,              & \text{otherwise} \\
\end{cases} dt
\end{equation}
where $t$ is a measure of the time that has passed since the time when the clutch was engaged $t_{\text{clutch}}$.

The slip threshold $F_{\text{slip}}$ of the clutch can depend on many factors. Naturally, it is expected to increase with increasing levels of $P_2$. It may also depend on the amount of material enclosed in the clutch $x$, the rate of the extension $\dot{x}$, the vacuum actuation pressure $P_1$ and the normal force between the surfaces of the inverting tube from the clutching pressure $P_2$. 

\subsection{Fabrication and Experimental Characterization}
\label{sec:exp_characterization}

\begin{figure}[!t]
\centering
\includegraphics[width=3.4in]{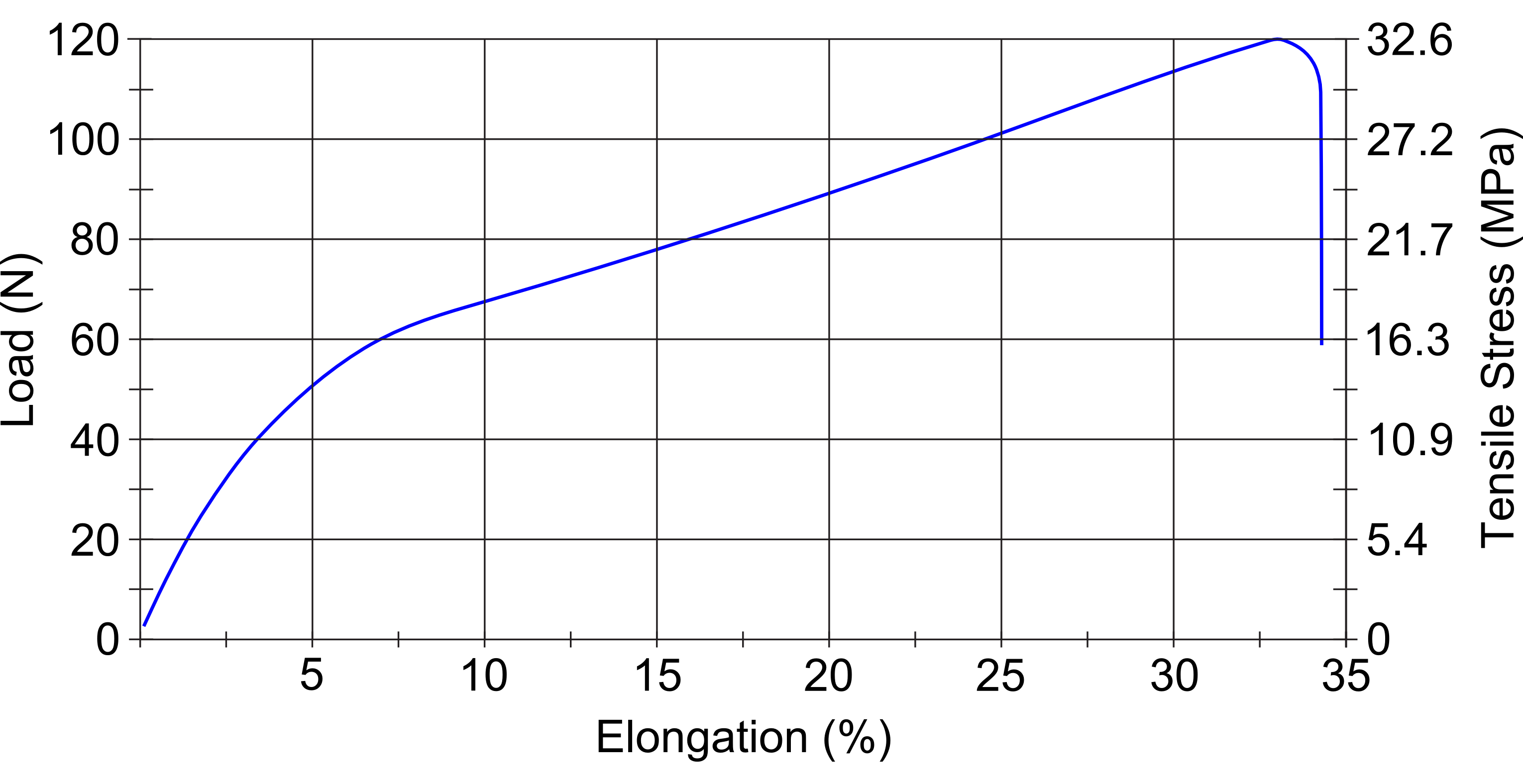}
\caption{A stress-strain curve for the inverting tube. A sample of the layflat polyethylene tubing was tested in tension. The material exhibited strain-hardening as it underwent plastic deformation.}
\label{fig:Materials}
\end{figure}

The InVACC prototypes tested in this work were fabricated using a tee-connection at the bottom of a length of reinforced PVC tubing (\unit[1.9]{cm} ID, McMaster Carr, 5393K45). 
The reinforced tubing was cut to \unit[26]{cm} beyond the stem of the internal barbed tee.
A short (\unit[5]{cm}) length of copper tubing  (\unit[1.9]{cm} OD, \unit[0.8]{mm} wall thickness) was inserted into the top of the reinforced tube.
The copper tube served as a connection point for the clutch and the inverting tube. 
The inverting tube was made from ``layflat'' polyethylene tubing with a measured width (half the circumference) of \unit[29]{mm} and a thickness of \unit[63.5]{$\mu$m} (LF1/250, Bayquest Packaging, UK).
The clutch was made with tubing that was twice as thick (\unit[127]{$\mu$m}) but otherwise the same (LF1/500).

\begin{figure}
    \centering
    \includegraphics[width = 3.4 in]{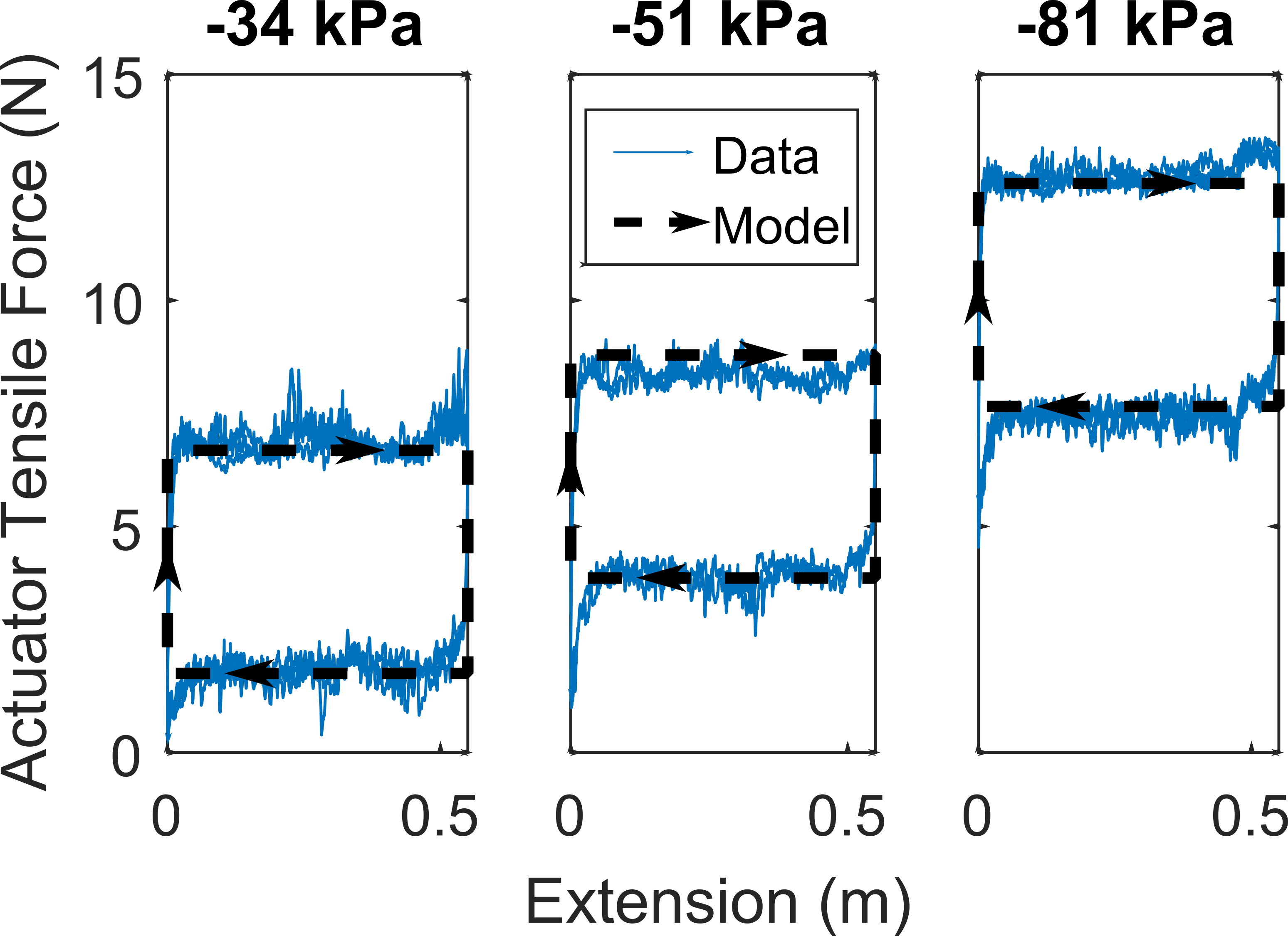}
    \caption{Shown are the actuator tensile force data for repeating cycles of strain across the entire stroke-length. The model (thick dashed line) is able to capture this behavior with an average RMS error of \unit[0.3]{N}. The actuation force increases linearly with the magnitude of the vacuum pressure applied to $P_1$. The losses are approximately constant and lead to an approximately \unit[5]{N} difference between the extending force (top line) and the retracting force (bottom line).  What appears as noise in the data are variations in forces from the random patterns of deformations that occur in the inverting tube.}
    \label{fig:ActuatorForceResults}
\end{figure}

The clutching tube was fed through the tee to one side, partially inverted and connected to the side connection.
The pressure on that side of the tee was thus connected to the space between the inside of the clutch tubing and the inverting tubing ($P_1$).
The other side of the tee was connected to the space between the outside of the clutch tubing and the inside of the reinforced tubing ($P_2$). Small sections of thin pneumatic tubing were placed in the membrane tubing within the tee to break the seal around the corners.

The device and material was characterized in tension on an Instron 5965 with a \unit[500]{N} dynamic force sensor (Instron 2580-05). For the device tests, the jaws of the testing machine were connected to \unit[60]{cm} of inverting tubing (measured from the end of the device). The tube was inverted into the device and the zero point for the extension measurements was set at \unit[1.5]{cm} from the end of the device. The nominal stroke lengths for the tests was \unit[55]{cm}. The inverting tube failed (pinhole leak) multiple times during testing. When this occurred, the inverting material was replaced. The data presented in this work are from tubes that were determined to be without failure. The vacuum pressure was drawn from a venturi vacuum ejector with a regulated pressure input.

\section{RESULTS}
\label{sec:results}

The device succeeded in switching between actuation and clutching modes. A video showing the operation of the device can be found at the following link {https://vimeo.com/272212723}. 

\subsection{Inverting-Tube Material Characterization}

To understand the behavior of the inverting tube under stress, the bare material was tested under strain (Fig.~\ref{fig:Materials}). The sample had a nominal cross-sectional area of approximately \unit[3.7]{mm$^2$}. The sample tested began to yield under a tensile load of about \unit[60]{N} (corresponding to a stress on the order of \unit[16]{MPa} with approximately \unit[7]{\%} strain). The ultimate tensile strength of the sample was \unit[120]{N} (stress of \unit[32.6]{MPa}) at approximately \unit[33]{\%} strain. 

\subsection{Actuator Characterization}

To characterize the force output of the actuator, the device was tested for three full-stroke cycles (\unit[0-55]{cm}) at various pressures and strain rates (Fig~\ref{fig:ActuatorForceResults}).
The actuator was tested at three levels of vacuum pressure applied to both $P_1$ and $P_2$, \unit[-34]{kPa}, \unit[-51]{kPa} and \unit[-81]{kPa}.
At the lowest-magnitude pressure, the actuator was tested at three strain rates: \unitfrac[100]{mm}{min}, \unitfrac[500]{mm}{min} and \unitfrac[2500]{mm}{min} (the maximum supported by the instrument). The forces were analyzed at extensions between \unit[5]{cm} and \unit[50]{cm} and averaged independently for extension and contraction (Table \ref{tab:actuator_force}).

\begin{figure}
    \centering
    \includegraphics[width = 2.5 in]{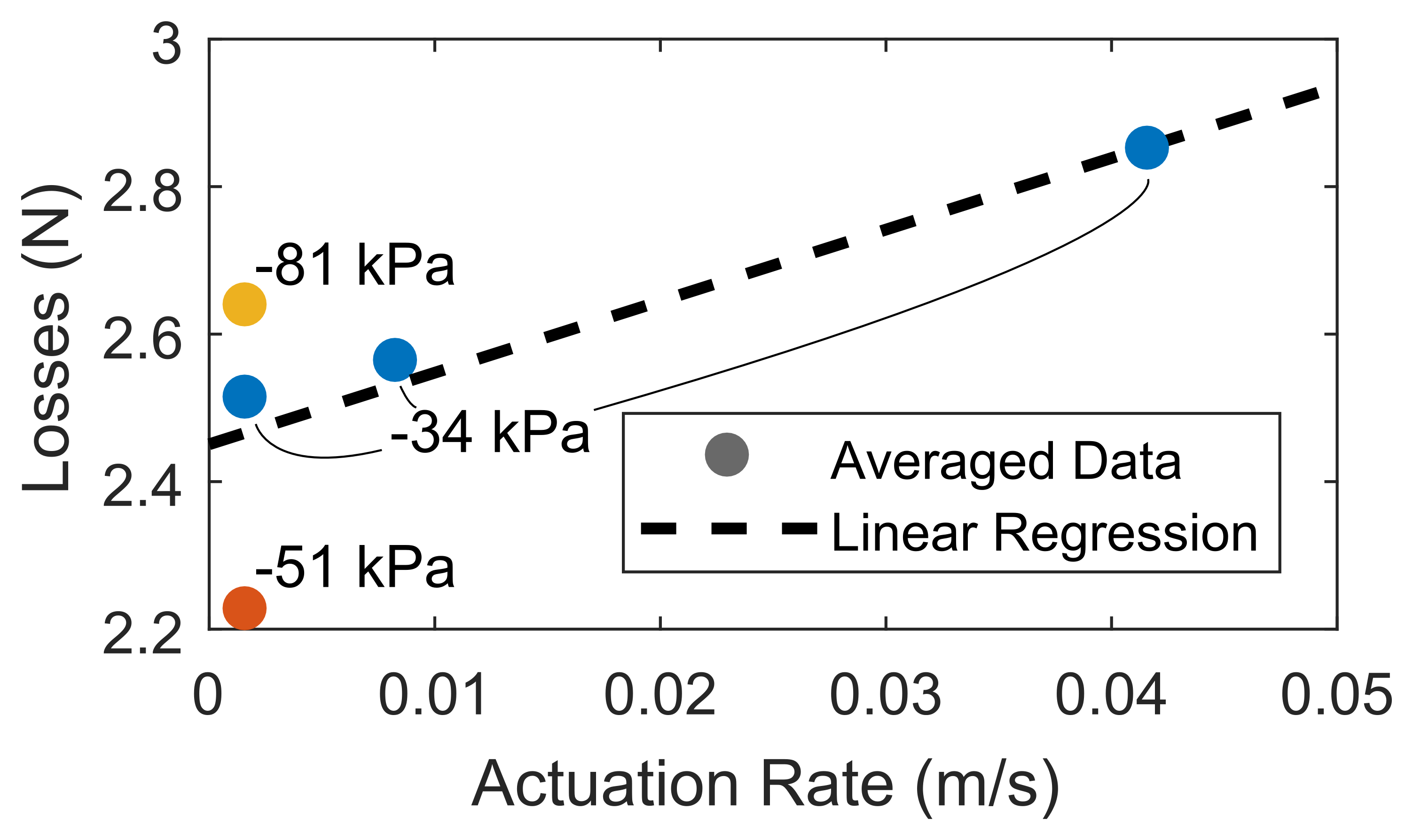}
    \caption{The average losses recorded during actuation for different pressures and actuation rates. The yield force loss was approximately \unit[2.5]{N}. There were also viscous losses that increased in magnitude with the actuation rate.}
    \label{fig:ActuatorLosses}
\end{figure}

The losses were calculated from half the difference between the averaged extension and contraction forces (Fig.~\ref{fig:ActuatorLosses}). A linear regression was applied to these data to identify the coefficients of Eq.~\eqref{eq:force_model_specific} (Table \ref{tab:actuator_force_coeficients}). Over the analyzed extension range, the average residual of the model-predicted force to the experimental data was \unit[0.3]{N}. The difference between the model-predicted force and the average force over the analyzed extension range was, on average, different by only \unit[2.6]{\%} of the data magnitude.

\begin{table}[h]
    \centering
     \caption{Identified Actuator Model Coefficients}
    \begin{tabular}{|c|l|l|}\hline
      Parameter   & Value & Note \\ \hline \hline
        $A_{\text{eff}}$ &  \unit[2.49]{cm$^2$} &  \unit[93]{\%} of inverting-tube cross section \\ \hline
        $F_{\text{yield}}$   &   \unit[2.45]{N} & Measured \unit[2.2]{N}-\unit[2.6]{N} at \unit[100]{mm/min} \\ \hline
        $\mu_{\text{visc}}$  &   \unit[9.70]{N/(m/s)} & Tested up to \unit[2500]{mm/min} \\
     \hline
    \end{tabular}   
    \label{tab:actuator_force_coeficients}
\end{table}

\begin{figure}
    \centering
    \includegraphics[width = 2.5 in]{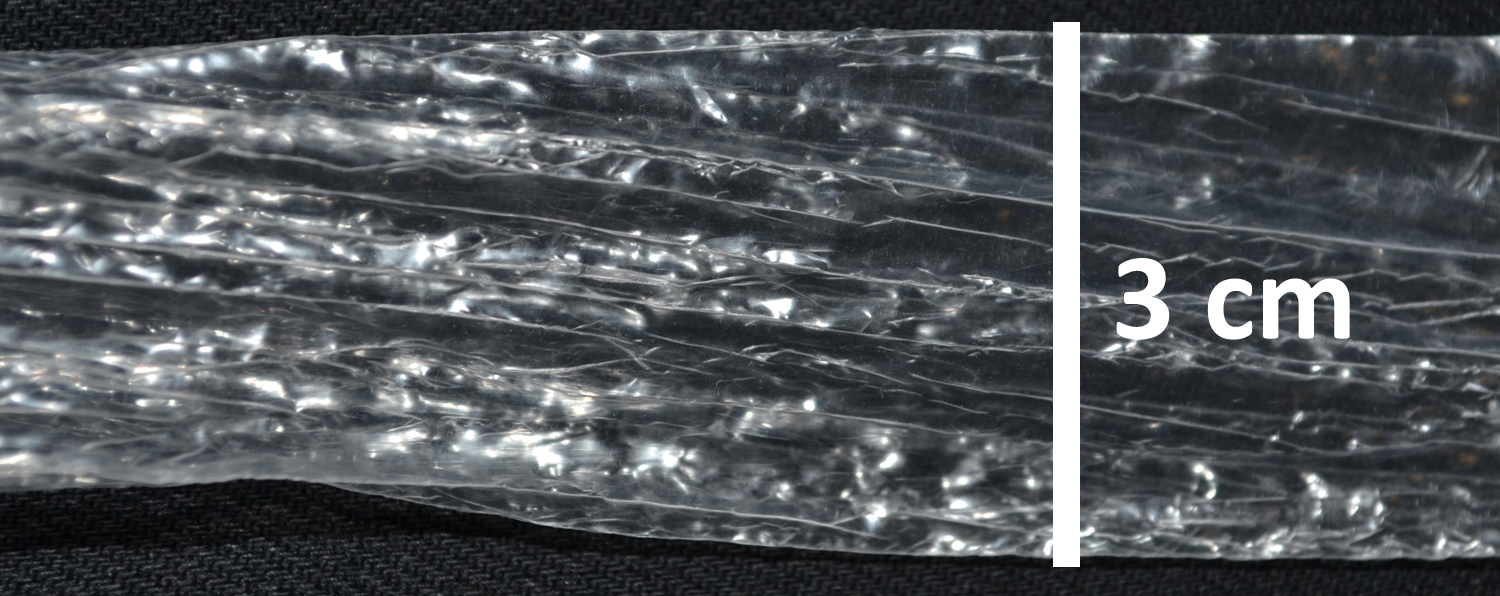}
    \caption{The inverting-tube material underwent substantial plastic deformation under repeated testing. The corresponding embrittlement of the material led to failure (leaking) after only a few cycles.}
    \label{fig:Wrinkled material}
\end{figure}

The inverting-tube material developed leaks after only a few actuation cycles and was repeatedly replaced during testing. Examining the material (Fig.~\ref{fig:Wrinkled material}) revealed stress-concentration points that had undergone repeated plastic deformation during the pressurized inversion.

\subsection{Clutch Characterization}

The clutch performance was characterized by forcibly extending the clutched device at a fixed strain rate (\unit[100]{mm/min}). A vacuum retraction pressure ($P_1 =$ \unit[-34]{kPa}) was applied to ensure that the tube everted properly (i.e. did not bunch up) during the forced extension. The clutch was tested at pressures ($P_2$) between \unit[10]{kPa} and \unit[50]{kPa} in \unit[10]{kPa} increments. 

\begin{figure}
    \centering
    \includegraphics[width = 2.5 in]{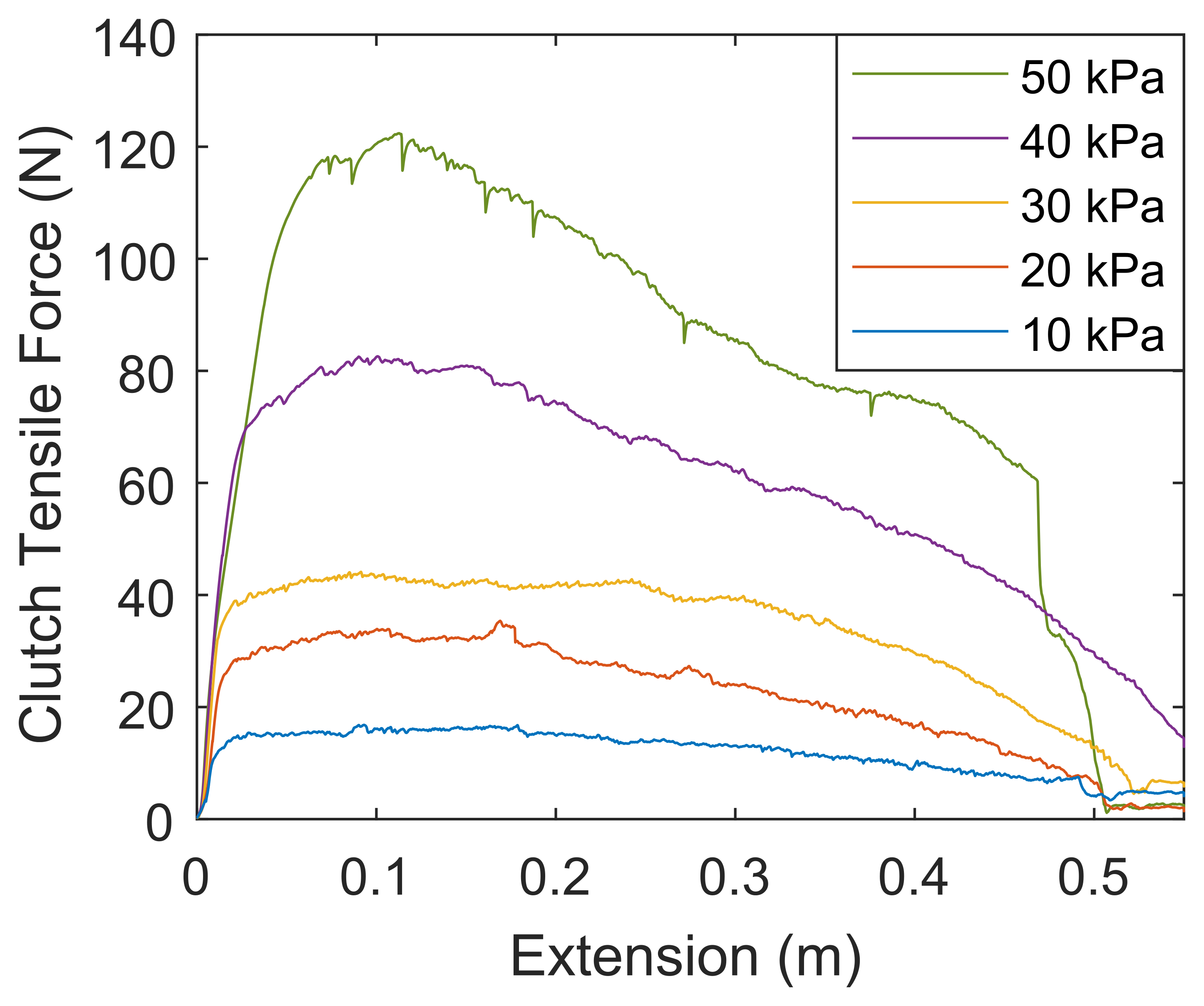}
    \caption{Shown are the forces recorded while forcing the device to extend with the clutch engaged. The tension develops gradually in the system from the elasticity of the tendon-like inverting tube. The tension peaks at low extensions and decreases as the inverting tube extends out from the internal clutching membrane.}
    \label{fig:ClutchForceResults}
\end{figure}

The slipping force $F_{\text{slip}}$ from clutch increased non-linearly with increasing clutching pressure (Fig.~\ref{fig:ClutchForceResults}, Table \ref{tab:ClutchPeaks}). Note that the extension data in Fig.~\ref{fig:ClutchForceResults} have been shifted left slightly so that zero extension corresponds to the point when \unit[0.25]{N} tension occurred. This shift compensates for the small plastic elongation that developed in the inverting tube over the course of the tests. 

\begin{table}[hp]
    \centering
    \caption{Peak Values of Slip Force with Clutch}
   \begin{tabular}{|c|c|c|}\hline
$P_2$ & Peak Tension & Peak Tension/Pressure \\
(kPa) & (N) & (N/kPa) \\ \hline \hline
10 & 17 & 1.68 \\ \hline
20 & 35 & 1.77 \\ \hline
30 & 44 & 1.47 \\ \hline
40 & 83 & 2.07 \\ \hline
50 & 122 & 2.45 \\ \hline
\end{tabular}
    \label{tab:ClutchPeaks}
\end{table}

\begin{table*}[]
    \centering
    \caption{Actuator Characterization Results}
    \begin{tabular}{|c|c|c|c|c|c|c|} \hline
 $P_1$ & Actuation Rate & Range & Model Force & Force Data, Mean (SD) & Residual(RMS) & \% Error \\
(kPa) & (mm/min) & \unit[0.5-50]{cm} & (N) & (N) & (N) & of Model to Data Mean \\ \hline \hline
\multirow{3}{*}{-34} & \multirow{3}{*}{100} & Extension & 6.68 & 6.83 (0.29) & 0.33 & -2.2 \\
& & Contraction & 1.75 & 1.81 (0.27) & 0.28 & -3.2 \\
& & Both & 4.22 & 4.32 (2.53) & 0.30 & -2.4 \\ \hline
\multirow{3}{*}{-34} & \multirow{3}{*}{500} & Extension & 6.75 & 6.88 (0.28) & 0.31 & -2.0 \\
& & Contraction & 1.69 & 1.76 (0.28) & 0.29 & -4.2 \\
& & Both & 4.22 & 4.32 (2.58) & 0.30 & -2.4 \\ \hline
\multirow{3}{*}{-34} & \multirow{3}{*}{2500} & Extension & 7.07 & 7.01 (0.33) & 0.33 & 0.8 \\
& & Contraction & 1.36 & 1.31 (0.35) & 0.36 & 3.7 \\
& & Both & 4.22 & 4.16 (2.87) & 0.34 & 1.3 \\ \hline
\multirow{3}{*}{-51} & \multirow{3}{*}{100} & Extension & 8.79 & 8.34 (0.23) & 0.51 & 5.4 \\
& & Contraction & 3.86 & 3.89 (0.26) & 0.26 & -0.8 \\
& & Both & 6.32 & 6.11 (2.24) & 0.40 & 3.5 \\ \hline
\multirow{3}{*}{-81} & \multirow{3}{*}{100} & Extension & 12.59 & 12.72 (0.22) & 0.26 & -1.1 \\
& & Contraction & 7.65 & 7.45 (0.24) & 0.31 & 2.7 \\
& & Both & 10.12 & 10.09 (2.65) & 0.29 & 0.3 \\ \hline \hline
& &      &       &              & \textbf{Average$\colon$} \unit[0.33]{N} & \textbf{Average$\colon$} \unit[2.6]{\%} \\
& &      &       &              &                                 & (Abs. values of Ext. and Contr.) \\ \hline
    \end{tabular}
    \label{tab:actuator_force}
\end{table*}

\section{DISCUSSION}
\label{sec:disc}

\begin{table*}[htp]
    \centering
    \caption{Comparison to Other Soft Linear Clutch technologies}
    \begin{tabular}{|c|c|c|c|c|c|c|c|} \hline
      \multirow{2}{*}{\textbf{Technology}}    & \multirow{2}{*}{\textbf{Example}}  & \multirow{2}{*}{\textbf{Switching}}  & \textbf{Retraction} & \textbf{Multi-layer}   &  \textbf{Demonstrated}     & \textbf{Surface Pressure}         & \textbf{Max.}    \\  & &   & \textbf{Force}      & \textbf{Compatible?} &  \textbf{Surface Pressure} & \textbf{Limit} & \textbf{Strain}   \\ \hline \hline
     Electrostatic & Diller et al. & Electronic, & Elastic,   & \multirow{2}{*}{yes}          & \multirow{2}{*}{\unit[22]{kPa}} & Dielectric breakdown,   & \multirow{2}{*}{\unit[100]{\%}} \\ 
      Adhesion     & 2016 \cite{diller_lightweight_2016} & Fast & Fixed   &          &   & e.g. \textbf{\unit[62]{kPa}} \cite{LuxPrintdatasheet}  & \\ \hline
      Vacuum         & Choi et al. & Fluidic, & Elastic,    & \multirow{2}{*}{yes}       &     \multirow{2}{*}{\unit[94.5]{kPa}}            & Ambient pressure,                    & \multirow{2}{*}{\unit[100]{\%}} \\ 
     Layer-Jamming  & 2018 \cite{choi_soft_2018} &  Slow & Fixed   &       &                 &  \textbf{\unit[100]{kPa}} at sea level       &  \\ \hline
     \multirow{2}{*}{\textbf{InVACC}}  & \multirow{2}{*}{W.~Felt} &  Fluidic, & \textbf{Vacuum,} & \multirow{2}{*}{\textbf{no}} & \multirow{2}{*}{\unit[50]{kPa}} & Clutching fluid pressure,  & \multirow{2}{*}{\textbf{\unit[200]{\%}}}  \\
       &  &  Slow  & \textbf{Controllable} &  &  & order of \textbf{\unit[500]{kPa}} is feasible  &   \\\hline
    \end{tabular}
    \label{tab:ClutchComparison}
\end{table*}

As predicted by the proposed model, the vacuum actuation force recorded in this work was approximately constant over the length of the actuator stroke. The average of the extending and contracting forces was \unit[93]{\%} of the theoretical maximum predicted by the model (for the same pressure). Overall, the simple three-parameter actuator model fit to the data captured the averaged actuation forces with an error of only \unit[2.6]{\%}.

Though the actuation force did not substantially change with the magnitude of the extension, an approximately \unit[5]{N} difference between the extension and contraction forces was observed. This difference is described by the proposed model as twice the ``yield force.'' It exists in part from the plastic deformations in the inverting-tube material during extension and retraction. These repeated deformations caused leaks to form in the inverting tube after only a few cycles.  A small ``viscous'' loss that scaled with the rate of actuation was also observed. This has been observed in other pressure-driven inverting tubes \cite{HawkesGrowth2017}. The actuators in this work were relatively short and tested in straight conditions. Long, curved paths could introduce more frictional losses into the system \cite{HawkesGrowth2017}.

The use of other inverting-tube materials could reduce the magnitude of the yield force and increase the fatigue life of the InVAC actuator. Other researchers have used nylon fabric hermetically sealed with either impregnated silicone \cite{UsevitchICRA2018} or thermoplastic polyurethane (TPU) \cite{PolygerinosPipeInspection} for (non-inverting) soft pneumatic actuators. Pleated Pneumatic Artificial Muscles have relied on Kevlar fabric with a thin polypropylene liner \cite{daerden1999conception}. Selection of the inverting-tube material could keep the desired frictional and stiffness characteristics of the clutch in mind.

The vacuum actuation forces of an InVAC are limited by the ambient pressure and the diameter of the inverting tube. The actuation force comes from the difference between the ambient pressure and the internal pressure on the two sides of the inverting tube. This difference cannot exceed the magnitude of the ambient pressure. As is apparent in Eq.~\eqref{eq:force_model_general}, the ideal InVAC tensile force is the product of the pressure difference with \textit{half} the cross-sectional area of the inverting tube. This creates a fundamental maximum for actuation force that can be achieved with an InVAC of a given diameter (see Eq.~\eqref{eq:force_max}, Table \ref{tab:max_actuation_forces}).

To support tensile forces greater than what could be achieved with vacuum actuation alone, the InVAC actuator was integrated with an internal positive-pressure clutch (InVACC). The slipping force of the clutch is adjustable through the pressure applied to the outside of the collapsing clutch membrane. The highest slipping force observed in this work was \unit[120]{N} with \unit[50]{kPa} of clutching pressure.

To achieve rapidly engaging and stiff clutching forces, there are several design factors to consider. The largest clutching force that can be achieved is determined by the material limits of the device. The inverting tendon must be able to bear the tensile load and the clutching membrane must be able to support the pressure. Within these limits, the slip threshold will increase with increased clutching pressure. When the clutch is engaged, a tensile force on the inverting tube will cause elastic elongation, even if the clutch does not slip. By using stiff membrane materials for the inverting tube and clutching membrane, this passive stretch can be minimized. Another factor to consider is the switching time of the clutch. To switch from disengaged to engaged, the clutching pressure $P_2$ must be increased from negative ($\leq P_1$) to positive. Using high-flow valves and limiting the expanded volume of the clutch can help increase the switching speed.

Compared to similar soft linear clutch technologies, the InVACC has several advantages (Table \ref{tab:ClutchComparison}) including high interface pressures, controllable retraction forces and high strains. Though an InVACC does not have multiple layers, it can be designed for positive clutching pressures much higher than the limits found in vacuum layer-jamming or electrostatic clutches.  An InVACC has a switchable and controllable retraction/actuation force. Other clutches use elastic forces which cannot be ``turned off'' and typically grow in magnitude as the device extends. An InVACC can extend to three times its contracted length. Clutches based on overlapping sheets can at best double their length. 

If they can be miniaturized, the properties of an InVACC clutch may be especially beneficial at small scales. At small scales, the flexibility and bending radius of an InVACC's reinforced tubing will be improved. At small scales, where surface area is limited, the relatively high interface pressure of an InVACC compares favorably to the limited pressures of other technologies. Some difficulties of scaling down the InVACC concept include keeping the inverting tube thin enough to be retracted and keeping the yield forces low enough that slack can be managed with vacuum.

\section{CONCLUSION}
The Inverting-tube Vacuum ACtuator with Clutch (InVACC) device proposed and demonstrated in this work is able to create controllable tensile forces by modulating the pressure within its two separate chambers. By subjecting both chambers to vacuum pressure, the inverting-tube actuator can perform both positive and negative work through its respective retraction and extension. The inverting-tube actuator can contract down to a \textit{third} of its extended length, much shorter than most contractile SPAs. 

By using a positive clutching pressure to collapse a membrane onto the inverting tube, the extension of the device can be restricted. This clutching mode allows the inverting-tube tendon to support much larger tensile loads than possible in its actuating mode.  Compared to similar linear clutches, the InVACC can: extend twice as far, control its retraction force, and potentially create higher frictional pressures.

The combined clutch/actuator concept can create a large range of forces--\textit{larger} forces than a vacuum actuator of its same size and \textit{smaller} forces than a similar clutch. This makes the InVACC a promising candidate to enable and inspire new kinds of robotic devices, especially wearable haptic systems.

\section*{ACKNOWLEDGMENT}

The author's salary during the time of this work was supported by the EU FLAG-ERA project RoboCom++. The author acknowledges discussions with Vivek Ramachandran that led to the inspiration for this work. Thanks to the Reconfigurable Robotics Laboratory (RRL) of EPFL for hosting this research.



\begin{thebibliography}{10}
\providecommand{\url}[1]{#1}
\csname url@samestyle\endcsname
\providecommand{\newblock}{\relax}
\providecommand{\bibinfo}[2]{#2}
\providecommand{\BIBentrySTDinterwordspacing}{\spaceskip=0pt\relax}
\providecommand{\BIBentryALTinterwordstretchfactor}{4}
\providecommand{\BIBentryALTinterwordspacing}{\spaceskip=\fontdimen2\font plus
\BIBentryALTinterwordstretchfactor\fontdimen3\font minus
  \fontdimen4\font\relax}
\providecommand{\BIBforeignlanguage}[2]{{%
\expandafter\ifx\csname l@#1\endcsname\relax
\typeout{** WARNING: IEEEtran.bst: No hyphenation pattern has been}%
\typeout{** loaded for the language `#1'. Using the pattern for}%
\typeout{** the default language instead.}%
\else
\language=\csname l@#1\endcsname
\fi
#2}}
\providecommand{\BIBdecl}{\relax}
\BIBdecl

\bibitem{daerden_pneumatic_2002}
F.~Daerden and D.~Lefeber, ``Pneumatic {Artificial} {Muscles}: actuators for
  robotics and automation,'' \emph{European journal of mechanical and
  environmental engineering}, 2002.

\bibitem{Chou1996}
C.-P. Chou and B.~Hannaford, ``Measurement and modeling of {McKibben} pneumatic
  artificial muscles,'' \emph{IEEE Transactions on robotics and automation},
  vol.~12, no.~1, pp. 90--102, 1996.

\bibitem{hawkes_300_strain_2016}
E.~W. Hawkes, D.~L. Christensen, and A.~M. Okamura, ``Design and implementation
  of a 300\% strain soft artificial muscle,'' in \emph{2016 {IEEE}
  {International} {Conference} on {Robotics} and {Automation} ({ICRA})}, May
  2016, pp. 4022--4029.

\bibitem{hammond_pneumatic_2017}
Z.~M. Hammond, N.~S. Usevitch, E.~W. Hawkes, and S.~Follmer, ``Pneumatic {Reel}
  {Actuator}: {Design}, modeling, and implementation,'' in \emph{2017 {IEEE}
  {International} {Conference} on {Robotics} and {Automation} ({ICRA})}, May
  2017, pp. 626--633.

\bibitem{HawkesGrowth2017}
\BIBentryALTinterwordspacing
E.~W. Hawkes, L.~H. Blumenschein, J.~D. Greer, and A.~M. Okamura, ``A soft
  robot that navigates its environment through growth,'' \emph{Science
  Robotics}, vol.~2, no.~8, 2017. [Online]. Available:
  \url{http://robotics.sciencemag.org/content/2/8/eaan3028}
\BIBentrySTDinterwordspacing

\bibitem{Belforte_textile}
\BIBentryALTinterwordspacing
G.~Belforte, G.~Eula, A.~Ivanov, and A.~L. Visan, ``Bellows textile muscle,''
  \emph{The Journal of The Textile Institute}, vol. 105, no.~3, pp. 356--364,
  2014. [Online]. Available: \url{https://doi.org/10.1080/00405000.2013.840414}
\BIBentrySTDinterwordspacing

\bibitem{RodrigueVacuumBellows}
J.~Lee and H.~Rodrigue, ``Origami-based vacuum pneumatic artificial muscle with
  large contraction ratio,'' \emph{(Under Review)}.

\bibitem{UsevitchICRA2018}
N.~S. Usevitch, A.~M. Okamura, and E.~W. Hawkes, ``{APAM}: Antagonistic
  pneumatic artificial muscle,'' in \emph{IEEE International Conference on
  Robotics and Automation, in press}, 2018.

\bibitem{feltVacuumBellows}
W.~Felt, M.~A. Robertson, and J.~Paik, ``Modeling vacuum bellows soft pneumatic
  actuators with optimal mechanical performance,'' in \emph{IEEE-RAS
  International Conference on Soft Robotics (RoboSoft)}, 2018.

\bibitem{choi_soft_2018}
I.~Choi, N.~Corson, L.~Peiros, E.~W. Hawkes, S.~Keller, and S.~Follmer, ``A
  {Soft}, {Controllable}, {High} {Force} {Density} {Linear} {Brake} {Utilizing}
  {Layer} {Jamming},'' \emph{IEEE Robotics and Automation Letters}, vol.~3,
  no.~1, pp. 450--457, Jan. 2018.

\bibitem{collins_reducing_2015}
\BIBentryALTinterwordspacing
S.~H. Collins, M.~B. Wiggin, and G.~S. Sawicki,
  ``\BIBforeignlanguage{en}{Reducing the energy cost of human walking using an
  unpowered exoskeleton},'' \emph{\BIBforeignlanguage{en}{Nature}}, vol. 522,
  no. 7555, pp. 212--215, Jun. 2015. [Online]. Available:
  \url{https://www.nature.com/articles/nature14288}
\BIBentrySTDinterwordspacing

\bibitem{diller_lightweight_2016}
S.~Diller, C.~Majidi, and S.~H. Collins, ``A lightweight, low-power
  electroadhesive clutch and spring for exoskeleton actuation,'' in \emph{2016
  {IEEE} {International} {Conference} on {Robotics} and {Automation} ({ICRA})},
  May 2016, pp. 682--689.

\bibitem{IROS_rolling_diaphram}
J.~P. {Whitney}, M.~F. {Glisson}, E.~L. {Brockmeyer}, and J.~K. {Hodgins}, ``A
  low-friction passive fluid transmission and fluid-tendon soft actuator,'' in
  \emph{2014 IEEE/RSJ International Conference on Intelligent Robots and
  Systems}, Sep. 2014, pp. 2801--2808.

\bibitem{kim_novel_2013}
Y.~J. Kim, S.~Cheng, S.~Kim, and K.~Iagnemma, ``A {Novel} {Layer} {Jamming}
  {Mechanism} {With} {Tunable} {Stiffness} {Capability} for {Minimally}
  {Invasive} {Surgery},'' \emph{IEEE Transactions on Robotics}, vol.~29, no.~4,
  pp. 1031--1042, Aug. 2013.

\bibitem{langer_stiffening_2018}
\BIBentryALTinterwordspacing
M.~Langer, E.~Amanov, and J.~Burgner-Kahrs, ``Stiffening {Sheaths} for
  {Continuum} {Robots},'' \emph{Soft Robotics}, Mar. 2018. [Online]. Available:
  \url{https://www.liebertpub.com/doi/10.1089/soro.2017.0060}
\BIBentrySTDinterwordspacing

\bibitem{LuxPrintdatasheet}
\BIBentryALTinterwordspacing
{DuPont}, ``{{LuxPrint} 8153 Electroluminescent Material},'' 2009. [Online].
  Available: \url{{\small
  http://www.dupont.com/content/dam/dupont/products-and-services/electronic-and-electrical-materials/documents/prodlib/8153.pdf}}
\BIBentrySTDinterwordspacing

\bibitem{PolygerinosPipeInspection}
W.~Adams, S.~Sridar, C.~M. Thalman, B.~Copenhaver, H.~Elsaad, and
  P.~Polygerinos, ``Water pipe robot utilizing soft inflatable actuators,'' in
  \emph{IEEE-RAS International Conference on Soft Robotics (RoboSoft)}, 2018.

\bibitem{daerden1999conception}
F.~Daerden, ``Conception and realization of pleated pneumatic artificial
  muscles and their use as compliant actuation elements,'' \emph{Vrije
  Universiteit Brussel}, p. 176, 1999.

\end{thebibliography}
\end{document}